\documentclass{article}

\PassOptionsToPackage{numbers, compress}{natbib}

\usepackage{files/arxiv_single_column/arxiv_single_column}

\usepackage[utf8]{inputenc} %
\usepackage[T1]{fontenc}    %
\usepackage{hyperref}       %
\usepackage{url}            %
\usepackage{booktabs}       %
\usepackage{amsfonts}       %
\usepackage{nicefrac}       %
\usepackage{microtype}      %
\usepackage{xcolor}         %
\usepackage{multirow}
\usepackage{tabularx}
\usepackage{subcaption}
\usepackage{graphicx}
\usepackage{enumitem}

\definecolor{darkblue}{rgb}{0,0 ,0.6}
\hypersetup{
    colorlinks = true,
    citecolor = darkblue,
    linkcolor = {black},
    urlcolor  = {black},
}

\usepackage{mathtools} 
\usepackage{microtype} 
\usepackage{nicefrac}       
\usepackage{pifont}   
\usepackage{siunitx}
\usepackage{threeparttable}  

\usepackage[labelfont=bf, font=small, margin=0pt]{caption}

\title{MMSFormer: Multimodal Transformer for Material and Semantic Segmentation}

\author{Md Kaykobad Reza$^1$, Ashley Prater-Bennette$^2$, M. Salman Asif$^1$ \\
$^1$ University of California Riverside, CA 92508, USA \\
$^2$ Air Force Research Laboratory, Rome, NY 13441, USA \\
\texttt{mreza025@ucr.edu, ashley.prater-bennette@us.af.mil, sasif@ucr.edu}
}

\begin{document}

\maketitle

\begin{abstract}
Leveraging information across diverse modalities is known to enhance performance on multimodal segmentation tasks. However, effectively fusing information from different modalities remains challenging due to the unique characteristics of each modality. In this paper, we propose a novel fusion strategy that can effectively fuse information from different modality combinations. We also propose a new model named {\underline M}ulti-{\underline M}odal {\underline S}egmentation Trans{\underline{Former}} (MMSFormer) that incorporates the proposed fusion strategy to perform multimodal material and semantic segmentation tasks. MMSFormer outperforms current state-of-the-art models on three different datasets. As we begin with only one input modality, performance improves progressively as additional modalities are incorporated, showcasing the effectiveness of the fusion block in combining useful information from diverse input modalities. Ablation studies show that different modules in the fusion block are crucial for overall model performance. Furthermore, our ablation studies also highlight the capacity of different input modalities to improve performance in the identification of different types of materials. The code and pretrained models will be made available at \url{https://github.com/csiplab/MMSFormer}.  
\end{abstract}

\section{Introduction}
Image segmentation \cite{cheng2001color, minaee2021image} methods assign one class label to each pixel in an image. The segmentation map can be used for holistic understanding of objects or context of the scene. Image segmentation can be further divided into different types; examples include semantic segmentation \cite{guo2018semseg, wang2018semseg2}, instance segmentation \cite{gu2022instanceseg2, hafiz2020instanceseg}, panoptic segmentation \cite{elharrouss2021panoptic2, kirillov2019panoptic} and material segmentation \cite{Liang2022MCubeS, upchurch2022dense}. Each of these segmentation tasks are designed to address specific challenges and applications. 

Multimodal image segmentation \cite{guo2019deep2, zhang2021deep} aims to enhance the accuracy and completeness of the task by leveraging diverse sources of information, and potentially leading to a more robust understanding of complex scenes. In contrast to single-modal segmentation \cite{minaee2021imagesegsinglemod}, the multimodal approach \cite{zhang2021imagesegmm} is more complex due to the necessity of effectively integrating heterogeneous data from different modalities. Key challenges arise from variations in data quality and attributes, distinct traits of each modality, and need to create models capable of accurately and coherently segmenting with the fused information.

Most of the existing multimodal segmentation methods are designed to work with specific modality pairs, such as RGB-Depth \cite{chen2020sa-gate, hazirbas2017fusenet, hu2019acnet}, RGB-Thermal \cite{li2023RSFNet, liang2023eaef, sun2019rtfnet}, and RGB-Lidar \cite{li2023mseg3d, prakash2021TransFuser, zhao2021lifseg}. As they are designed for specific modality combinations, most of them generally do not work well with modality combinations different from the ones used in the original design. Recently, CMX \cite{zhang2023cmx} introduced a technique to fuse information from RGB and one other supplementary modality, but it is incapable of fusing more than two modalities at the same time. Some recent models have proposed techniques to fuse more than two modalities \cite{broedermann2023hrfuser, Liang2022MCubeS, zhang2023CMNext}. However, they either use very complex fusion strategies \cite{broedermann2023hrfuser, zhang2023cmx} or require additional information like semantic labels \cite{Liang2022MCubeS} for performing underlying tasks. 

\begin{figure*}[t]
     \centering
     \includegraphics[width=0.91\linewidth]{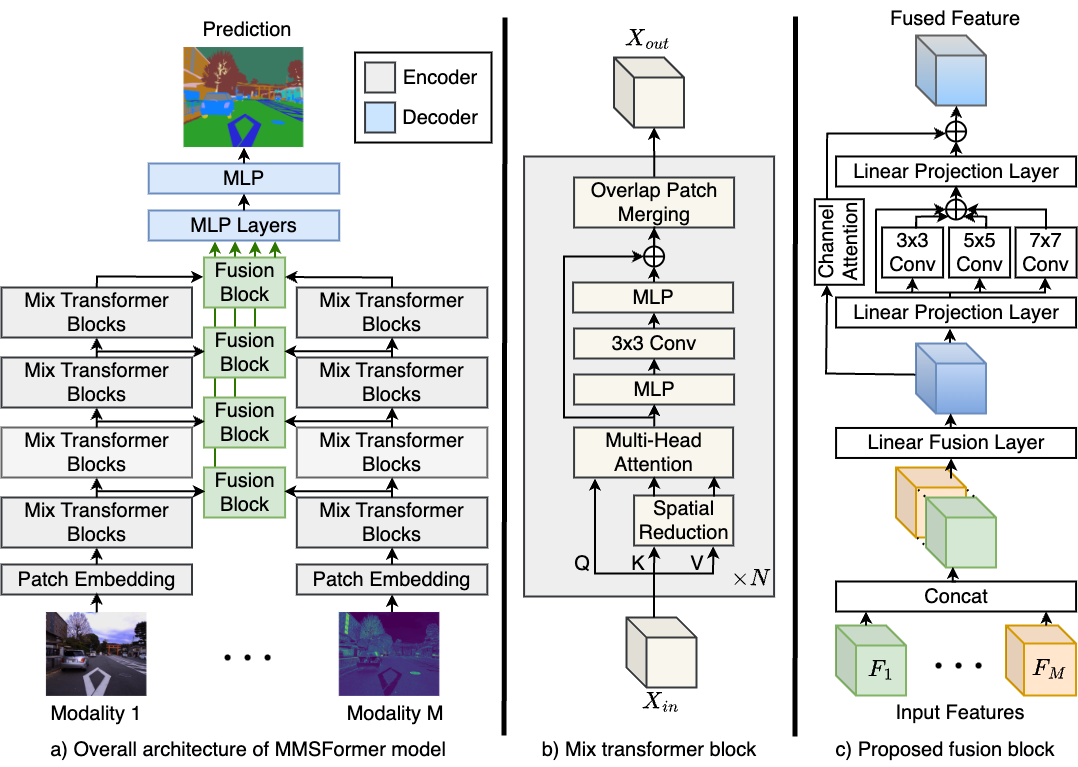}
    \caption{(a) Overall architecture of MMSFormer model. Each image passes through a modality-specific encoder where we extract hierarchical features. Then we fuse the extracted features using the proposed fusion block and pass the fused features to the decoder for predicting the segmentation map. (b) Illustration of the mix transformer \cite{xie2021segformer} block. Each block applies a spatial reduction before applying multi-head attention to reduce computational cost. 
    (c) Proposed multimodal fusion block. We first concatenate all the features along the channel dimension and pass it through linear fusion layer to fuse them. Then the feature tensor is fed to linear projection and parallel convolution layers to capture multi-scale features. We use Squeeze and Excitation block \cite{hu2019squeezeandexcitation} as channel attention in the residual connection to dynamically re-calibrate the features along the channel dimension.}
    \label{fig:model-architecture}
\end{figure*}

In this paper, we propose a novel fusion block that can fuse information from diverse combination of modalities. We also propose a new model for multimodal material and semantic segmentation tasks that we call MMSFormer. Our model uses transformer based encoders \cite{xie2021segformer} to capture hierarchical features from different modalities, fuses the extracted features with our novel fusion block and utilizes MLP decoder to perform multimodal material and semantic segmentation. In particular, our proposed fusion block uses parallel convolutions to capture multi-scale features, channel attention to re-calibrate features along the channel dimension and linear layer to combine information across multiple modalities. Such a design provides a simple and computationally efficient fusion block that can handle an arbitrary number of input modalities and combine information effectively from different modality combinations. We compare our fusion block with some of the existing fusion methods in terms of number of parameters and GFLOPs in Table~\ref{tab:gflop-param-comparison}.

To evaluate our proposed MMSFormer and fusion block, we focus on multimodal material segmentation on MCubeS \cite{Liang2022MCubeS} dataset and multimodal semantic segmentation on FMB \cite{liu2023segmif} and PST900 \cite{Shivakumar2020PST900} datasets. MCubeS dataset consists of four different modalities: RGB, angle of linear polarization (AoLP), degree of linear polarization (DoLP) and near-infrared (NIR). FMB dataset includes RGB and infrared modalities, while PST900 dataset comprises RGB and thermal modalities. We show the overall and per-class performance comparison in Table~\ref{tab:mcubes-performance-comparison-with-sota}-\ref{tab:pst-comparison-with-sota} for these datasets.
A series of experiments highlight the ability of the proposed fusion block to effectively combine features from different modality combinations, resulting in superior performance compared to current state-of-the-art methods. Ablation studies show that different input modalities assist in identifying different types of material classes as shown in Table~\ref{tab:mcubes-per-class-iou-for-modality-combination}. Furthermore, as we add new input modalities, overall performance increases graduallyhighlighting the ability of the fusion block to incorporate useful information from new modalities. We summarize the results in Table~\ref{tab:fmb-per-class-iou-comparison} and \ref{tab:mcubes-modality-combination-performance-comparison} for FMB and MCubeS datasets respectively.

Main contributions of this paper can be summarized as follows. 
\begin{itemize}
    \item We propose a new multimodal segmentation model called MMSFormer. The model incorporates a novel fusion block that can fuse information from arbitrary (heterogeneous) combinations of modalities.
    \item Our model achieves new state-of-the-art performance on three different datasets. Furthermore, our method achieves better performance for all modality combinations compared to the current leading models.
    \item A series of ablation studies show that each module on the fusion block has an important contribution towards the overall model performance and each input modality assists in identifying specific material classes.
\end{itemize}
Rest of the paper is structured as follows. Section~\ref{sec:related-work} presents a brief review of related work. We describe our model and fusion block in detail in Section~\ref{sec:proposed-model}. Section~\ref{sec:experiment} presents experimental results and ablation studies on multimodal material and semantic segmentation tasks with qualitative and quantitative analysis. 
\section{Related Work}
\label{sec:related-work}

Image segmentation has witnessed significant evolution, spurred by advancements in machine learning and computational capabilities. A significant improvement in this evolution came with the inception of fully convolutional networks (FCNs) \cite{long2015FCN, wu2019fastfcn}, which enabled pixel-wise predictions through the utilization of hierarchical features within convolutional neural networks (CNNs). This led to the development of a variety of CNN based models for different images segmentation tasks. U-Net \cite{ronneberger2015unet} is one such model that utilizes skip-connections between the lower-resolution and corresponding higher-resolution feature maps. DeepLabV3+ \cite{chen2018deeplabv3+} introduced dilated convolutions (atrous convolutions) into the encoder allowing the expansion of the receptive field without increasing computational complexity significantly. PSPNet \cite{zhao2017pspnet} introduced global context modules that enable the model to gather information from a wide range of spatial scales, essentially integrating both local and global context into the segmentation process. 

Recently, Transformer based models have proven to be very effective in handling complex image segmentation tasks. Some of the notable transformer-based models are  Pyramid Vision Transformer (PVT) \cite{wang2021pvt}, SegFormer \cite{xie2021segformer}, and Mask2Former \cite{cheng2022mask2former}. PVT \cite{wang2021pvt} utilizes transformer based design for various computer vision tasks. SegFormer \cite{xie2021segformer} utilizes efficient self-attention and lightweight MLP decoder for simple and efficient semantic segmentation. Mask2Former \cite{cheng2022mask2former} uses masked-attention along with pixel decoder and transformer decoder for any segmentation task. Their success demonstrates the capacity of these models to provide state-of-the-art solutions in various segmentation tasks.

In the context of multimodal image segmentation, fusion of data from diverse sources \cite{zhang2021deep} has gained traction as a means to extract richer information and improve accuracy. A variety of models and fusion strategies have been proposed for RGB-Depth segmentation tasks. FuseNet \cite{hazirbas2017fusenet} model integrates depth feature maps into RGB feature maps, while SA-Gate \cite{chen2020sa-gate} employs Separation-and-Aggregation Gating to mutually filter and recalibrate RGB and depth modalities before fusion. Attention Complementary Module has been proposed by ACNet \cite{hu2019acnet} that extracts weighted RGB and depth features for fusion. The domain of RGB-Thermal image segmentation has also gained prominence. Recent models include RTFNet \cite{sun2019rtfnet} that  achieves fusion through elementwise addition of thermal features with RGB, RSFNet \cite{li2023RSFNet} proposing Residual Spatial Fusion module to blend RGB and Thermal modalities, and EAEFNet \cite{liang2023eaef} utilizing attention interaction and attention complement mechanisms to merge RGB and Thermal features. A number of methods also focus on fusing RGB-Lidar data that include TransFuser \cite{prakash2021TransFuser} that employs Transformer blocks, whereas LIF-Seg \cite{zhao2021lifseg} relies on coarse feature extraction, offset learning, and refinement for effective fusion.

While the previously mentioned studies focus on specific pairs of modalities, some recent research has demonstrated promising results in the fusion of arbitrary modalities. CMX \cite{zhang2023cmx} introduces cross-modal feature rectification and fusion modules to merge RGB features with supplementary modalities. For multimodal material segmentation, MCubeSNet \cite{Liang2022MCubeS} model is proposed, which can seamlessly integrate four different modalities to enhance segmentation accuracy. In the context of arbitrary modal semantic segmentation, CMNeXt \cite{zhang2023CMNext} introduces Self-Query Hub and Parallel Pooling Mixer modules, offering a versatile approach for fusing diverse modalities. Additionally, HRFuser \cite{broedermann2023hrfuser} employs multi-window cross-attention to fuse different modalities at various resolutions, thereby enriching model performance.

Though some of these models can fuse different modalities, they either use very complex fusion strategies \cite{broedermann2023hrfuser, zhang2023cmx, zhang2023CMNext} or requires additional information \cite{Liang2022MCubeS} to perform underlying task. We aim to design a simple fusion module that can handle arbitrary number of input modalities and able to effectively fuse information from diverse modality combinations.
\section{Proposed Model}
\label{sec:proposed-model}
The overall architecture of our proposed MMSFormer model and the fusion block is shown in Figure~\ref{fig:model-architecture}. The model has three modules: (1) Modality specific encoder; (2) Multimodal fusion block; and (3) Shared MLP decoder. We use mix transformer \cite{xie2021segformer} as the encoder of our model. We choose mix transformer for various reasons. First, it can provide hierarchical features without positional encoding. Second, it uses spatial reduction before attention that reduces the number of parameters significantly \cite{wang2021pvt, xie2021segformer}. Third, it also works well with simple and lightweight MLP decoder \cite{xie2021segformer}. 

\subsection{Overall Model Architecture}
Our overall model architecture is shown in Figure~\ref{fig:model-architecture}a. Assume we have $M$ distinct modalities. Given a set of modalities as input, each modality-specific encoder captures distinctive features from each input modality by mapping the corresponding image into modality-specific hierarchical features as
\begin{equation}\label{eq:encoder}
    F_m = \text{Encoder}_m(I_m),%
\end{equation}
where $I_m \in \mathbb{R}^{H \times W \times 3}$ represents input image for modality $m \in \{1, 2, \ldots, M\}$ and Encoder$_m(\cdot)$ denotes the encoder for that modality. The encoder %
generates four feature maps at $\{\frac{1}{4}, \frac{1}{8}, \frac{1}{16}, \frac{1}{32}\}$ of the input image resolution. We represent them as $F_m = \{F_m^1, F_m^2, F_m^3, F_m^4\}$. For simplicity we denote the shape of the feature map for the $i^{th}$ encoder stage as $(H_i \times W_i \times C_i)$ where $i \in \{1, 2, 3, 4\}$.     

We use four separate fusion blocks, one corresponding to each encoder stage, to fuse the features from each stage of the encoder. We pass the extracted features $F_m^i$ for all modalities to the $i^{th}$ fusion block as  
\begin{equation}\label{eq:fusion}
    F^i = \text{FusionBlock}^i(\{F_m^i\}_m).
\end{equation}
Each fusion block fuses the features extracted from all the modalities to generate a combined feature representation $F = \{F^1, F^2, F^3, F^4\}$, where $F^i$ denotes the fused feature at $i^{th}$ stage. 
Finally, we pass the combined features $F$ to the MLP decoder \cite{xie2021segformer} to predict the segmentation labels.

\subsection{Modality Specific Encoder}
We use mix transformer encoder \cite{xie2021segformer} to capture hierarchical features from the input modalities. Each input image $I_m$ goes through patch embedding layer where it is divided into $4 \times 4$ patches following \cite{xie2021segformer} and then fed to the mix transformer blocks. The design of mix transformer block is shown in Figure~\ref{fig:model-architecture}b. We denote the input to any mix transformer block as $X_{in} \in \mathbb{R}^{H_i \times W_i \times C_i}$ that is reshaped to $N_i \times C_i$ (with  $N_i = H_iW_i$) and used as query $Q$, key $K$, and value $V$. 

To reduce the computational overhead, spatial reduction is applied following \cite{wang2021pvt} using a reduction ratio $R$. $K$ and $V$ are first reshaped into $\frac{N_i}{R} \times C_iR$ matrices and then mapped to $\frac{N_i}{R} \times C_i$ matrices via linear projection. 
A standard multi-head self-attention (MHSA) maps $Q, K,V$ to intermediate features as
\begin{equation}
    \begin{split}
        \text{MHSA}(Q, K, V) &= \text{Concat}(\text{head}_1, \dots , \text{head}_h)W^O, \\
        \text{head}_j &= \text{Attention}(QW_j^Q, KW_j^K, VW_j^V),
    \end{split}
\end{equation}
where $h$ represents the number of attention heads, $W_j^Q \in \mathbb{R}^{C_i \times d_K}$, $W_j^K \in \mathbb{R}^{C_i \times d_K}$, $W_j^V \in \mathbb{R}^{C_i \times d_V}$, and $W^O \in \mathbb{R}^{hd_V \times C_i}$ are the projection matrices, $d_K, d_V$ represent dimensions of $K, V$, respectively. We can formulate the Attention function as
\begin{equation}
    \text{Attention}(Q, K, V) = \text{Softmax}(\frac{QK^T}{\sqrt{d_{K}}})V,
\end{equation}
where $Q, K,$ and $V$ are the input query, key, and value matrices. 
MHSA is followed by a mixer layer (with two MLP and one $3 \times 3$ convolution layer). The convolution layer provides sufficient positional encoding into the transformer encoder for optimal segmentation performance \cite{xie2021segformer}. This layer can be written as 
\begin{equation}
    \begin{split}
        \hat{X}_{in} &= \text{MHSA}(Q, K, V), \\
        X_{out} &= \text{MLP}(\text{GELU}(\text{Conv}_{3 \times 3}(\text{MLP}(\hat{X}_{in})))) + \hat{X}_{in},
    \end{split}
\end{equation}
Finally, overlap patch merging is applied to $X_{out}$ following \cite{xie2021segformer} to generate the final output. %

\subsection{Multimodal Fusion Block}
After extracting hierarchical features, we fuse them using our proposed fusion block. The fusion block shown in Figure~\ref{fig:model-architecture}c is responsible for fusing the features extracted from the modality specific encoders. We have one fusion block for each of the four encoder stages. For the $i^{th}$ fusion block, let us assume the input feature maps are given as $F_m^i \in \mathbb{R}^{H_i \times W_i \times C_i}~\forall m \in \{1, 2, \dots, M\}$. First, we concatenate the feature maps from $M$ modalities along the channel dimension to get the combined feature map $F^i \in \mathbb{R}^{H_i \times W_i \times MC_i}$. Then we pass the features through a linear fusion layer that combines the features and reduces the channel dimension to $C_i$. Let us denote the resulting features as $\hat{F}^i \in \mathbb{R}^{H_i \times W_i \times C_i}$. We represent the operation as 
\begin{equation}\label{eq:cat-mlp}
    \hat{F}^i = \text{Linear}(F_1^i || \cdots || F_M^i).
\end{equation}
Here $||$ represents concatenation of features along the channel dimension and the linear layer takes an $MC_i$ dimensional input and generates a $C_i$ dimensional output.  

After the linear fusion layer, we added a module for capturing and mixing multi-scale features. The module consists of two linear projection layers having parallel convolution layers in between them. First we apply a linear transformation on $\hat{F}^i$ along the channel dimension by passing it through the first linear projection layer. It refines and tunes the features from different channels. Then we apply $3\times 3$, $5\times 5$, and $7\times 7$ convolutions to effectively capture diverse features across different spatial contexts. By employing convolutions with different sizes, the fusion block can attend to local patterns as well as capture larger spatial structures, thereby enhancing its ability to extract meaningful features from the input data. Finally we apply another linear transformation along the channel dimension using the second linear project layer to consolidate the information captured by the parallel convolutions, promoting feature consistency and enhancing the discriminative power of the fused features. These steps can be performed as 
\begin{equation}\label{eq:mix-mlp-1}
    \Tilde{F^i} = \text{Linear}(\hat{F^i}), 
\end{equation}
\begin{equation}\label{eq:mix-mlp-2}
    F^i = \text{Linear}(\Tilde{F^i} + \sum_{k \in \{3, 5, 7\}} \text{Conv}_{k \times k}(\Tilde{F^i})). 
\end{equation}

We found that using 3 parallel convolution layers with sizes $3\times 3$, $5\times 5$, and $7\times 7$ provide optimal performance. Increasing the convolution kernel size reduces performance which we show in Table~\ref{tab:ablation-study-on-fusion-block}. As larger kernels reduce performance, we did not add more than 3 parallel convolutions in our model.

We apply Squeeze-and-Excitation block \cite{hu2019squeezeandexcitation} as channel attention in the residual connection. The final fused feature can be represented as 
\begin{equation}\label{eq:residual}
    F^i = \text{ChannelAttention}(\hat{F^i}) + F^i.
\end{equation}
Channel attention re-calibrates interdependence between channels and allows the model to select the most relevant features or channels while suppressing less important ones \cite{hu2019squeezeandexcitation}. This leads to more effective feature representations and thus better performance on the underlying task.

\subsection{Shared MLP Decoder}
The fused features generated from all the 4 fusion blocks are sent to the shared MLP decoder. We use the deocder design proposed in \cite{xie2021segformer}. The decoder shown in Figure~\ref{fig:model-architecture}a can be represented as the following equations:
\begin{equation}
    \begin{split}
        F^i &= \text{Linear}(F^i),~~ \forall i \in \{1, 2, 3, 4\} \\
        F^i &= \text{Upsample}(F^i),~~ \forall i \in \{1, 2, 3, 4\} \\
        F &= \text{Linear}(F^1|| \cdots ||F^4), \\
        P &= \text{Linear}(F). 
    \end{split}
\end{equation}
The first linear layers take the fused features of different shapes and generate features having the same channel dimension. Then the features are up-sampled to $\frac{1}{4}^{th}$ of the original input shape, concatenated along the channel dimension and passed through another linear layer to generate the final fused feature $F$. Finally $F$ is passed through the last linear layer to generate the predicted segmentation map $P$.

\section{Experiments and Results}
\label{sec:experiment}
We evaluated our model and proposed fusion block on multiple datasets and with different modality combinations for multimodal semantic and material segmentation tasks. We also compared our methods with existing baseline methods both qualitatively and quantitatively. We report results from already published works whenever possible. $\ast$ indicates that we have used the code and pretrained models from the papers to generate the results.

\begin{table}[t]
    \begin{minipage}{0.48\linewidth}
    \centering
    \captionof{table}{Performance comparison on Multimodal Material Segmentation (MCubeS) dataset \cite{Liang2022MCubeS}. Here A, D, and N represent angle of linear polarization (AoLP), degree of linear polarization (DoLP), and near-infrared (NIR) respectively.}
    \begin{tabular}{lcc}
        \toprule
        Method                                 & Modalities & \% mIoU \\
        \midrule
        \midrule
        DRConv  \cite{chen2021DRConv}            & RGB-A-D-N  & 34.63 \\
        DDF  \cite{zhou2021DDF}                  & RGB-A-D-N  & 36.16 \\
        TransFuser  \cite{prakash2021TransFuser} & RGB-A-D-N  & 37.66 \\
        DeepLabv3+  \cite{chen2018deeplabv3+}    & RGB-A-D-N  & 38.13 \\
        MMTM  \cite{joze2020mmtm}                & RGB-A-D-N  & 39.71 \\
        FuseNet  \cite{hazirbas2017fusenet}      & RGB-A-D-N  & 40.58 \\
        MCubeSNet  \cite{Liang2022MCubeS}        & RGB-A-D-N  & 42.46 \\
        CBAM  \cite{woo2018cbam}        & RGB-A-D-N  & 51.32 \\
        CMNeXt    \cite{zhang2023CMNext}  & RGB-A-D-N  & 51.54 \\  
        \textbf{MMSFormer (Ours)}                & RGB-A-D-N  & \textbf{53.11} \\
        \bottomrule
    \end{tabular}
    \label{tab:mcubes-performance-comparison-with-sota}
\end{minipage}
    \hfill
    \begin{minipage}{0.48\linewidth}
    \centering
    \captionof{table}{Performance comparison on FBM \cite{liu2023segmif} dataset. We show performance for different methods from already published works. %
    }
    \begin{tabular}{lcc}
    \toprule
    \multicolumn{1}{c}{Methods} & Modalities      & \% mIoU \\
    \midrule
    \midrule
    CBAM \cite{woo2018cbam}           & RGB-Infrared & 50.1    \\
    GMNet \cite{zhou2021GMNet}        & RGB-Infrared & 49.2    \\
    LASNet \cite{Li2023LASNet}        & RGB-Infrared & 42.5    \\
    EGFNet \cite{dong2023EGFNet}      & RGB-Infrared & 47.3    \\
    FEANet \cite{deng2021FEANet}      & RGB-Infrared & 46.8    \\
    DIDFuse \cite{Zhao2020DIDFuse}    & RGB-Infrared & 50.6    \\
    ReCoNet \cite{huang2022Reconet}   & RGB-Infrared & 50.9    \\
    U2Fusion \cite{xu2022U2Fusion}    & RGB-Infrared & 47.9    \\
    TarDAL \cite{liu2022tardal}       & RGB-Infrared & 48.1    \\
    SegMiF \cite{liu2023segmif}       & RGB-Infrared & 54.8    \\
    \textbf{MMSFormer (Ours)}         & RGB-Infrared & \textbf{61.7} \\
    \bottomrule
    \end{tabular}
    \label{tab:fmb-comparison-with-sota}
\end{minipage}
\end{table}

\subsection{Datasets}
{\bf Multimodal material segmentation (MCubeS) dataset} \cite{Liang2022MCubeS} contains 500 sets of images from 42 street scenes having four modalities: RGB, angle of linear polarization (AoLP), degree of linear polarization (DoLP), and near-infrared (NIR). It provides annotated ground truth labels for both material and semantic segmentation and divided into training set with 302 image sets, validation set with 96 image sets, and test set with 102 image sets. This dataset has 20 class labels corresponding to different materials. 

{\bf FMB dataset} 
\cite{liu2023segmif} is a new and challenging dataset with 1500 pairs of calibrated RGB-Infrared image pairs. The training and test set contains 1220 and 280 image pairs respectively. The dataset covers a wide range of scenes under different lighting and weather conditions (Tyndall effect, rain, fog, and strong light). It also provides per pixel ground truth annotation for 14 different classes.

{\bf PST900 dataset} \cite{Shivakumar2020PST900}
contains 894 pairs of synchronized RGB-Thermal image pairs. The dataset is divided into training and test sets with per pixel ground truth annotation for five different classes.

\subsection{Implementation Details}
To ensure a fair comparison with prior models, we followed the same data preprocessing and augmentation strategies employed in previous studies \cite{Liang2022MCubeS, zhang2023CMNext, liu2023segmif}. We used the Mix-Transformer (MiT) \cite{xie2021segformer} encoder pretrained on the ImageNet \cite{deng2009ImageNet} dataset as the backbone for our model to extract features from different modalities. Each modality has a separate encoder. We used a shared MLP decoder introduced in SegFormer \cite{xie2021segformer} and used random initialization for it. We trained and evaluated all our models using two NVIDIA RTX 2080Ti GPUs and used PyTorch for model development. 

We utilized a polynomial learning rate scheduler with a power of 0.9 to dynamically adjust the learning rate during training. The first 10 epochs were designated as warm-up epochs with a learning rate of 0.1 times the original rate. For loss computation, we used the cross-entropy loss function. Optimization was performed using the AdamW \cite{loshchilov2017adamw} optimizer with an epsilon value of $10^{-8}$ and weight decay set to 0.01. For CBAM \cite{woo2018cbam}, we use the same encoder, decoder and hyperparameters used in our experiments and replace our fusion block with CBAM\footnote{https://github.com/luuuyi/CBAM.PyTorch} module. To be specific, after extracting the feature maps from each input modality using the modality specific encoders, we add them (sum them up) and pass the combined feature map to the CBAM module.

\subsection{Performance Comparison with Existing Methods}
We conducted a rigorous performance evaluation of our model compared to established baseline models for three datasets. The comprehensive results are summarized in Tables~\ref{tab:mcubes-performance-comparison-with-sota}--\ref{tab:mcubes-modality-combination-performance-comparison}. We report results for CBAM from our experiments. Other results are taken from published literature.

\textbf{Results on MCubeS Dataset.} Table~\ref{tab:mcubes-performance-comparison-with-sota} shows the overall performance comparison between our model and existing baseline models for MCubeS dataset. Our model achieves a mean intersection-over-union (mIoU) of 53.11\%, surpassing the current state-of-the-art model. It shows 1.57\% improvement over CMNeXt \cite{zhang2023CMNext}, 1.79\% improvement over CBAM \cite{woo2018cbam} and 10.65\% improvement over MCubeSNet \cite{Liang2022MCubeS} models. To further analyze the performance of our model, we conducted a per-class IoU analysis and presented in Table~\ref{tab:mcubes-per-class-iou-comparison-with-sota}. Our model performs better in detecting most of the material classes compared to the current state-of-the-art models. Notably, our model demonstrates a substantial improvement in the detection of plastic (+3.7\%), fabric (+3.1\%), asphalt (+2.3\%), and cobblestone (2.3\%) classes while maintaining competitive or better performance in other classes. This led to the overall better performance and sets new state-of-the-art for this dataset.

\begin{figure}[t]
  \centering
    \begin{subfigure}[b]{1.0\textwidth}
        \centering
        \includegraphics[width=\textwidth]{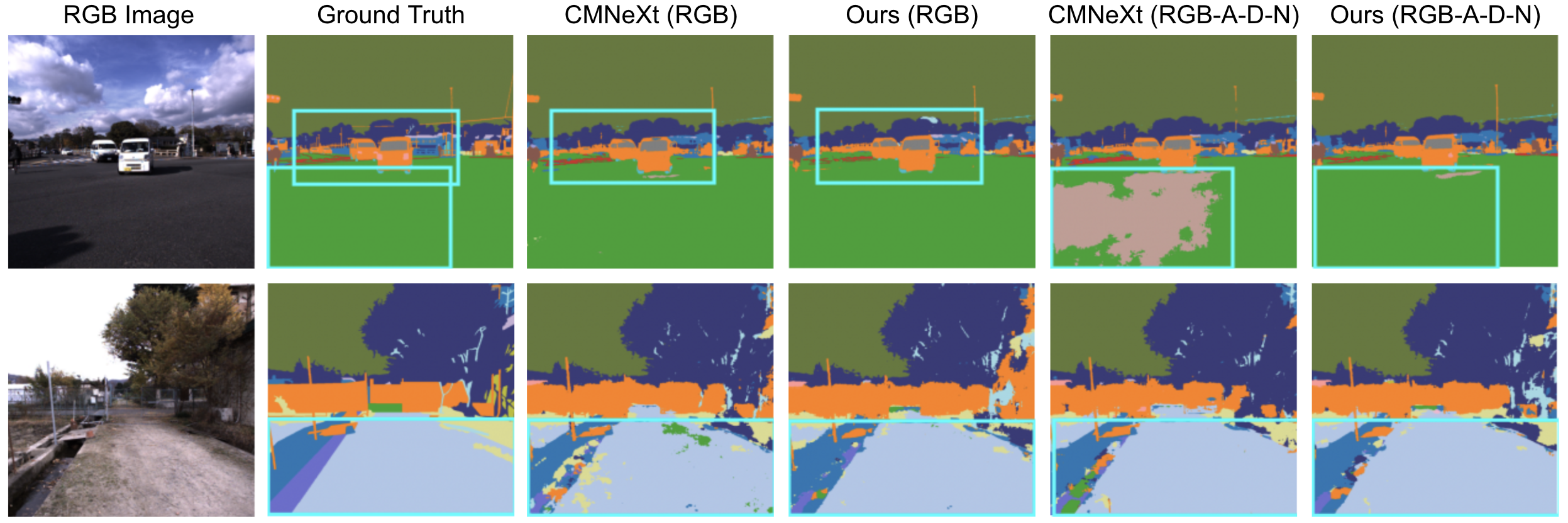}
        \caption{Visualization of predictions on MCubeS dataset}
        \label{fig:vis-mcubes-sota}
    \end{subfigure}
    \hfill
    \\[3pt]
    \begin{subfigure}[b]{1.0\textwidth}
        \centering
        \includegraphics[width=\textwidth]{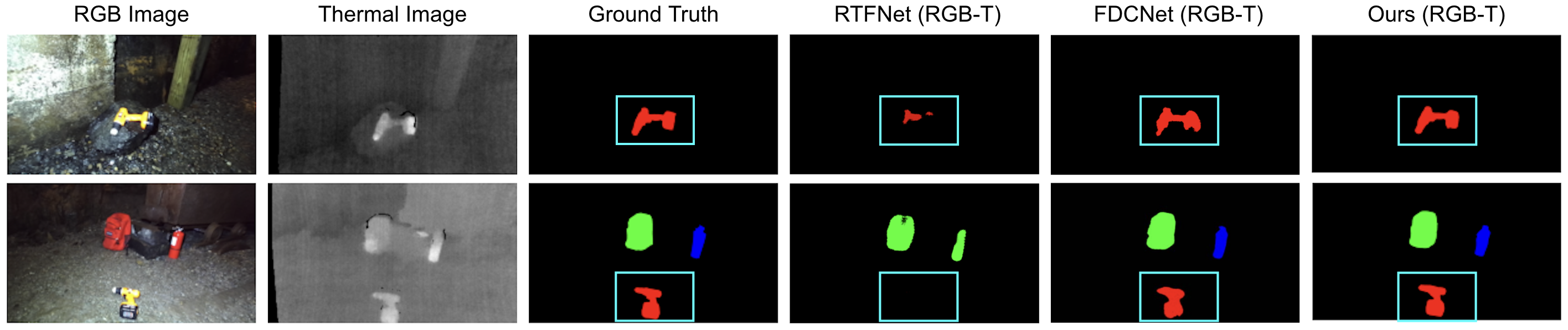}
        \caption{Visualization of predictions on PST900 dataset}
        \label{fig:vis-pst-sota}
    \end{subfigure} 
  \caption{Visualization of predictions on MCubeS and PST900 datasets. Figure~\ref{fig:vis-mcubes-sota} shows RGB and all modalities (RGB-A-D-N) prediction from CMNeXt \cite{zhang2023CMNext} and our model on MCubeS dataset. For brevity, we only show the RGB image and ground truth material segmentation maps along with the predictions. Figure~\ref{fig:vis-pst-sota} shows predictions from RTFNet \cite{sun2019rtfnet}, FDCNet \cite{zhao2023FDCNet} and our model for RGB-thermal input modalities on PST900 dataset. Our model shows better predictions on both of the datasets.}
  \label{fig:vis-prediction-sota}
\end{figure}

\textbf{Results on FMB Dataset.} Performance Comparison for FMB dataset is shown on Table~\ref{tab:fmb-comparison-with-sota}. Our model shows a significant improvement of 6.9\% mIoU compared to the current state-of-the-art model. Per-class IoU analysis for this dataset is shown on Table~\ref{tab:fmb-per-class-iou-comparison}. For a fair comparison, we only compare the performance on 8 classes (out of 14) that are published in literature. T-Lamp and T-Sign represent Traffic Lamp and Traffic Sign, respectively. Our model shows an overall performance improvement of 6.7\% mIoU for RGB only predictions compared to the most recent SegMiF \cite{liu2023segmif} model. Alongside this, our model also shows superior performance in detecting all of the classes except for the truck class for both RGB only and RGB-Infrared semantic segmentation tasks. Performance on RGB-Infrared input modalities is much better than RGB only performance for all the classes, which demonstrates the ability of the fusion block to effectively fuse information from the input modalities.

\textbf{Results on PST900 Dataset.} We also tested our model on PST900 \cite{Shivakumar2020PST900} dataset and summarized the result in Table~\ref{tab:pst-comparison-with-sota}. Experiments show that our model outperforms existing baseline models for RGB-Thermal semantic segmentation on this dataset. It outperforms the most recent CACFNet \cite{Zhou2023CACFNet} model by 0.89\% mIoU. Our model also shows better performance in detecting 3 out of the 5 classes available in the dataset and competitive performance in other two classes. 

\subsection{Performance Comparison for Incremental Modality Integration}
A critical aspect of this work involves evaluating the effectiveness of our proposed fusion block in combining valuable information from diverse modalities. To analyze this effect, we trained our model with various combinations of modalities on the MCubeS dataset. The results are presented in Table~\ref{tab:mcubes-modality-combination-performance-comparison}. Our model exclusively trained on RGB data provided an mIoU score of 50.44\%, which is 2.28\% grater than the current state-of-the-art model. We observe progressive improvement in performance as we incorporated additional modalities: AoLP, DoLP, and NIR. The integration led to incremental performance gains, with the mIoU increasing from 50.44\% to 51.30\%, then to 52.03\%, and ultimately reaching to 53.11\%. These findings serve as a compelling evidence that our fusion approach effectively leverages and fuses valuable information from different combination of modalities, resulting in a notable enhancement in segmentation performance. 

Furthermore, our model consistently outperforms the current state-of-the-art benchmark across all modality combinations. This consistent superiority underscores the robustness and versatility of our fusion block, demonstrating its ability to adapt and excel regardless of the specific modality combination provided. 

\begin{table*}[t]
    \centering
    \caption{Per-class \% IoU comparison on Multimodal Material Segmentation (MCubeS) \cite{Liang2022MCubeS} dataset. Our proposed MMSFormer model shows better performance in detecting most of the classes compared to the current state-of-the-art models. $\ast$ indicates that the code and pretrained model from the authors were used to generate the results.}
    \setlength{\tabcolsep}{2.0pt}
    \resizebox{\textwidth}{!}{
        \begin{tabular}{l|cccccccccccccccccccc|c}
            \toprule
            Methods &
            \rotatebox[origin=c]{90}{Asphalt} &
            \rotatebox[origin=c]{90}{Concrete} &
            \rotatebox[origin=c]{90}{Metal} &
            \rotatebox[origin=c]{90}{Road marking} &
            \rotatebox[origin=c]{90}{Fabric} &
            \rotatebox[origin=c]{90}{Glass} &
            \rotatebox[origin=c]{90}{Plaster} &
            \rotatebox[origin=c]{90}{Plastic} &
            \rotatebox[origin=c]{90}{Rubber} &
            \rotatebox[origin=c]{90}{Sand} &
            \rotatebox[origin=c]{90}{Gravel} &
            \rotatebox[origin=c]{90}{Ceramic} &
            \rotatebox[origin=c]{90}{Cobblestone} &
            \rotatebox[origin=c]{90}{Brick} &
            \rotatebox[origin=c]{90}{Grass} &
            \rotatebox[origin=c]{90}{Wood} &
            \rotatebox[origin=c]{90}{Leaf} &
            \rotatebox[origin=c]{90}{Water} &
            \rotatebox[origin=c]{90}{Human} &
            \rotatebox[origin=c]{90}{Sky} &
            \rotatebox[origin=c]{90}{Mean} \\ 
            \midrule
            \midrule
            
            MCubeSNet \cite{Liang2022MCubeS} &
            85.7 &
            42.6 &
            47.0 &
            59.2 &
            12.5 &
            44.3 &
            3.0 &
            10.6 &
            12.7 &
            66.8 &
            67.1 &
            27.8 &
            65.8 &
            36.8 &
            54.8 &
            39.4 &
            73.0 &
            13.3 &
            0.0 &
            94.8 &
            42.9 \\

            CBAM \cite{woo2018cbam} &
            85.7 &
            47.7 &
            55.4 &
            70.4 &
            27.6 &
            54.7 &
            \textbf{0.9} &
            30.9 &
            26.5 &
            61.6 &
            63.0 &
            28.0 &
            71.1 &
            41.8 &
            58.6 &
            47.4 &
            76.7 &
            \textbf{56.3} &
            25.9 &
            96.5 &
            51.3 \\
            
            CMNeXt \cite{zhang2023CMNext} $\ast$ &
            84.3 &
            44.9 &
            53.9 &
            \textbf{74.5} &
            32.3 &
            54.0 &
            0.8 &
            28.3 &
            \textbf{29.7} &
            \textbf{67.7} &
            66.5 &
            27.7 &
            68.5 &
            42.9 &
            58.7 &
            \textbf{49.7} &
            75.4 &
            55.7 &
            18.9 &
            96.5 &
            51.5 \\
        
            \textbf{MMSFormer (Ours)} &
            \textbf{88.0} &
            \textbf{48.3} &
            \textbf{56.2} &
            72.2 &
            \textbf{35.4} &
            \textbf{54.9} &
            0.5 &
            \textbf{34.6} &
            29.4 &
            67.2 &
            \textbf{69.0} &
            \textbf{29.9} &
            \textbf{73.4} &
            \textbf{44.7} &
            \textbf{59.5} &
            47.8 &
            \textbf{77.1} &
            50.5 &
            \textbf{26.9} &
            \textbf{96.6} &
            \textbf{53.1} \\ 
            \bottomrule
        \end{tabular}
    }
    \label{tab:mcubes-per-class-iou-comparison-with-sota}
\end{table*}

\begin{table*}[t]
    \centering
    \caption{Per-class \% IoU comparison on FMB \cite{liu2023segmif} dataset for both RGB only and RGB-infrared modalities. We show the comparison for 8 classes (out of 14) that are published. T-Lamp and T-Sign stand for Traffic Lamp and Traffic Sign respectively. Our model outperforms all the methods for all the classes except for the truck class.}
    \setlength{\tabcolsep}{4.0pt}
    \resizebox{\textwidth}{!}{
        \begin{tabular}{lcccccccccc}
            \toprule
            \multicolumn{1}{c}{Methods} & Modalities & Car           & Person        & Truck & T-Lamp        & T-Sign        & Building      & Vegetation    & Pole          & \% mIoU          \\
            \midrule
            \midrule
            SegMiF \cite{liu2023segmif}     & RGB & 78.3 & 46.6 & \textbf{43.4} & 23.7 & 64.0 & 77.8 & 82.1 & 41.8 & 50.5 \\
            \textbf{MMSFormer (Ours)}       & RGB & \textbf{80.3} & \textbf{56.7} & 42.1  & \textbf{31.6} & \textbf{77.8} & \textbf{77.9} & \textbf{85.4} & \textbf{48.1} & \textbf{57.2} \\
            \midrule
            CBAM \cite{woo2018cbam}      & RGB-Infrared & 71.9 & 49.3 & 20.9          & 25.8 & 67.1 & 75.8 & 80.9 & 19.7 & 50.1 \\
            GMNet \cite{zhou2021GMNet}      & RGB-Infrared & 79.3 & 60.1 & 22.2          & 21.6 & 69.0 & 79.1 & 83.8 & 39.8 & 49.2 \\
            LASNet \cite{Li2023LASNet}      & RGB-Infrared & 72.6 & 48.6 & 14.8          & 2.9  & 59.0 & 75.4 & 81.6 & 36.7 & 42.5 \\
            EGFNet \cite{dong2023EGFNet}    & RGB-Infrared & 77.4 & 63.0 & 17.1          & 25.2 & 66.6 & 77.2 & 83.5 & 41.5 & 47.3 \\
            FEANet \cite{deng2021FEANet}    & RGB-Infrared & 73.9 & 60.7 & 32.3          & 13.5 & 55.6 & 79.4 & 81.2 & 36.8 & 46.8 \\
            DIDFuse \cite{Zhao2020DIDFuse}  & RGB-Infrared & 77.7 & 64.4 & 28.8          & 29.2 & 64.4 & 78.4 & 82.4 & 41.8 & 50.6 \\
            ReCoNet \cite{huang2022Reconet} & RGB-Infrared & 75.9 & 65.8 & 14.9          & 34.7 & 66.6 & 79.2 & 81.3 & 44.9 & 50.9 \\
            U2Fusion \cite{xu2022U2Fusion}  & RGB-Infrared & 76.6 & 61.9 & 14.4          & 28.3 & 68.9 & 78.8 & 82.2 & 42.2 & 47.9 \\
            TarDAL \cite{liu2022tardal}     & RGB-Infrared & 74.2 & 56.0 & 18.8          & 29.6 & 66.5 & 79.1 & 81.7 & 41.9 & 48.1 \\
            SegMiF \cite{liu2023segmif}     & RGB-Infrared & 78.3 & 65.4 & \textbf{47.3} & 43.1 & 74.8 & 82.0 & 85.0 & 49.8 & 54.8 \\
            \textbf{MMSFormer (Ours)}       & RGB-Infrared & \textbf{82.6} & \textbf{69.8} & 44.6  & \textbf{45.2} & \textbf{79.7} & \textbf{83.0} & \textbf{87.3} & \textbf{51.4} & \textbf{61.7} \\
            \bottomrule
        \end{tabular}
    }
    \label{tab:fmb-per-class-iou-comparison}
\end{table*}

\begin{table*}[t]
    \caption{Performance comparison on PST900 \cite{Shivakumar2020PST900} dataset. We show per-class \% IoU as well as \% mIoU for all the classes.}
    \setlength{\tabcolsep}{4.0pt}
    \resizebox{\textwidth}{!}{
        \begin{tabular}{llcccccc}
            \toprule
            \multicolumn{1}{c}{Methods} & \multicolumn{1}{c}{Modalities} & Background     & Fire-Extinguisher & Backpack       & Hand-Drill     & Survivor & \% mIoU           \\
            \midrule
            \midrule
            ACNet \cite{hu2019acnet}         & RGB-Thermal & 99.25 & 59.95          & 83.19 & 51.46 & 65.19          & 71.81 \\
            CCNet \cite{Huang2019CCNet}         & RGB-Thermal & 99.05 & 51.84          & 66.42 & 32.27 & 57.50          & 61.42 \\
            Efficient FCN \cite{liu2020EfficientFCN} & RGB-Thermal & 98.63 & 39.96          & 58.15 & 30.12 & 28.00          & 50.98 \\
            RTFNet \cite{sun2019rtfnet}       & RGB-Thermal & 99.02 & 51.93          & 74.17 & 7.07  & 70.11          & 60.46 \\
            PSTNet \cite{Shivakumar2020PST900}       & RGB-Thermal & 98.85 & 70.12          & 69.20 & 53.60 & 50.03          & 68.36 \\
            EGFNet \cite{zhou2022EGFNet1}       & RGB-Thermal & 99.26 & 71.29          & 83.05 & 64.67 & 74.30          & 78.51 \\
            MTANet \cite{zhou2023MTANet}       & RGB-Thermal & 99.33 & 64.95          & 87.50 & 62.05 & 79.14          & 78.60 \\
            MFFENet \cite{Zhou2022MFFENet}      & RGB-Thermal & 99.40 & 66.38          & 81.02 & 72.50 & 75.60          & 78.98 \\
            GMNet  \cite{zhou2021GMNet}       & RGB-Thermal & 99.44 & 73.79          & 83.82 & 85.17 & 78.36          & 84.12 \\
            CGFNet  \cite{wang2022CGFNet}      & RGB-Thermal & 99.30 & 71.71          & 82.00 & 59.72 & 77.42          & 78.03 \\
            GCNet  \cite{Liu2022GCNet}       & RGB-Thermal & 99.35 & 77.68          & 79.37 & 82.92 & 73.58          & 82.58 \\
            GEBNet \cite{Dong2022GEBNet}       & RGB-Thermal & 99.39 & 73.07          & 85.93 & 67.14 & 80.21          & 81.15 \\
            GCGLNet \cite{GONG2023GCGLNet}      & RGB-Thermal & 99.39 & 77.57          & 81.01 & 81.90 & 76.31          & 83.24 \\
            DHFNet \cite{Cai2023DHFNet}       & RGB-Thermal & 99.44 & 78.15          & 87.38 & 71.18 & 74.81          & 82.19 \\
            MDRNet+  \cite{Zhao2023MDRNet+}     & RGB-Thermal & 99.07 & 63.04          & 76.27 & 63.47 & 71.26          & 74.62 \\
            FDCNet \cite{zhao2023FDCNet}     & RGB-Thermal & 99.15 & 71.52          & 72.17 & 70.36 & 72.36          & 77.11 \\
            CBAM \cite{woo2018cbam}     & RGB-Thermal & 99.43 & 73.81          & 82.75 & 80.00 & 69.60          & 81.12 \\
            EGFNet \cite{dong2023EGFNet}        & RGB-Thermal & 99.55 & 79.97          & 90.62 & 76.08 & \textbf{80.88} & 85.42 \\
            CACFNet \cite{Zhou2023CACFNet}      & RGB-Thermal & 99.57 & \textbf{82.08} & 89.49 & 80.90 & 80.76          & 86.56 \\
            \textbf{MMSFormer (Ours)}   & RGB-Thermal                    & \textbf{99.60} & 81.45             & \textbf{89.86} & \textbf{89.65} & 76.68    & \textbf{87.45} \\
            \bottomrule
        \end{tabular}
    }
    \label{tab:pst-comparison-with-sota}
\end{table*}

\begin{table}[t]
    \centering
    \caption{Performance comparison (\% mIoU) on Multimodal Material Segmentation (MCubeS) \cite{Liang2022MCubeS} dataset for different modality combinations. Here A, D, and N represent angle of linear polarization (AoLP), degree of linear polarization (DoLP), and near-infrared (NIR) respectively.}
    \begin{tabular}{lcccc}
        \toprule
        Modalities & MCubeSNet \cite{Liang2022MCubeS} & CMNeXt \cite{zhang2023CMNext} & MMSFormer (Ours)        \\
        \midrule
        \midrule
        RGB        & 33.70     & 48.16  & \textbf{50.44} \\
        RGB-A      & 39.10     & 48.42  & \textbf{51.30} \\
        RGB-A-D    & 42.00     & 49.48  & \textbf{52.03} \\
        RGB-A-D-N  & 42.86     & 51.54  & \textbf{53.11} \\
        \bottomrule
    \end{tabular}
    \label{tab:mcubes-modality-combination-performance-comparison}
\end{table}

\begin{figure}[t]
  \centering
    \begin{subfigure}[b]{1.0\textwidth}
        \centering
        \includegraphics[width=\textwidth]{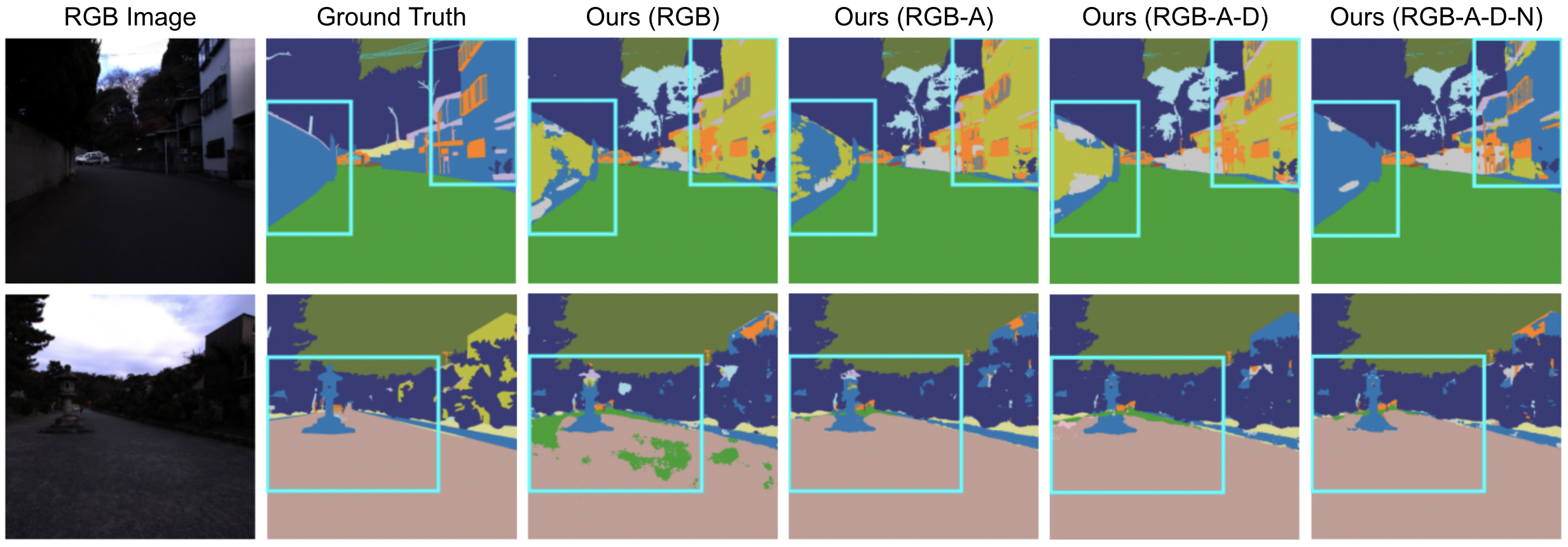}
        \caption{Visualization of predictions on MCubeS dataset for different modality combinations}
        \label{fig:vis-mcubes-modality-combination}
    \end{subfigure}
    \hfill
    \\[3pt]
    \begin{subfigure}[b]{1.0\textwidth}
        \centering
        \includegraphics[width=\textwidth]{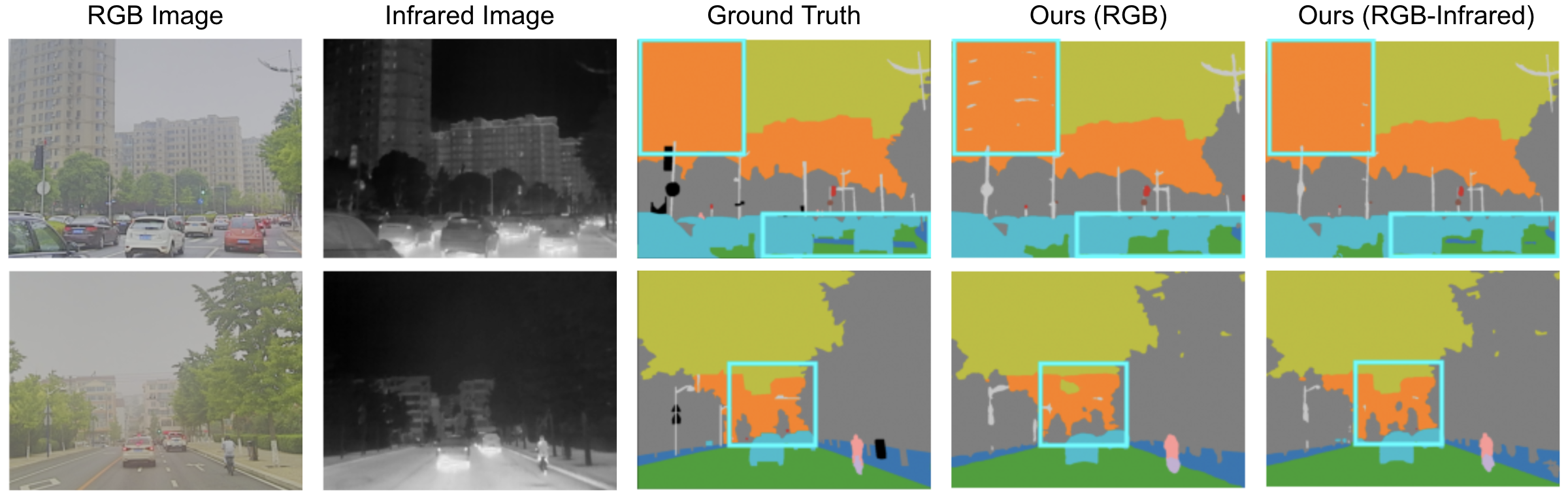}
        \caption{Visualization of predictions on FMB dataset for different modality combinations}
        \label{fig:vis-fmb-modality-combination}
    \end{subfigure} 
  \caption{Visualization of predicted segmentation maps for different modality combinations on MCubeS \cite{Liang2022MCubeS} and FMB \cite{liu2023segmif} datasets. Both figures show that prediction accuracy increases as we incrementally add new modalities. They also illustrate the fusion block's ability to effectively combine information from different modality combinations.}
  \label{fig:vis-prediction-modality-combination}
\end{figure}

\begin{table}[t]
    \centering
    \caption{Ablation study of the Fusion Block on FMB \cite{liu2023segmif} dataset. Both RGB and infrared input modalities were used during training and testing. The table shows the contribution of different modules in fusion block in overall model performance.}
    \begin{tabular}{lcc}
    \toprule
    \multicolumn{1}{c}{Structure}          & Parameter Count (M) & \% mIoU {\footnotesize \color{red} (Change)}                     \\
    \midrule
    \midrule
    MMSFormer                              & 61.26         & 61.68                              \\
    \midrule
    \hspace{2mm} - without channel attention            & 61.21          & 58.32 {\footnotesize \color{red} (-3.36)} \\
    \hspace{2mm} - without parallel convolutions        & 61.17          & 57.17 {\footnotesize \color{red} (-4.51)}                        \\
    \hspace{2mm} - with 3x3, 7x7 and 11x11 convolutions & 61.36          & 56.32 {\footnotesize \color{red} (-5.36)}                        \\
    \hspace{2mm} - only linear fusion                   & 59.57          & 52.43 {\footnotesize \color{red} (-9.25)}                        \\
    \bottomrule
    \end{tabular}
    \label{tab:ablation-study-on-fusion-block}
\end{table}

\begin{table*}[t]
    \centering
    \caption{Per class \% IoU comparison on Multimodal Material Segmentation (MCubeS) \cite{Liang2022MCubeS} dataset for different modality combinations. As we add modalities incrementally, overall performance increases gradually. This table also shows that specific modality combinations assist in identifying specific types of materials better.}
    \setlength{\tabcolsep}{2pt}
    \resizebox{\textwidth}{!}{
        \begin{tabular}{l|cccccccccccccccccccc|c}
            \toprule
            Modalities &
            \rotatebox[origin=c]{90}{Asphalt} &
            \rotatebox[origin=c]{90}{Concrete} &
            \rotatebox[origin=c]{90}{Metal} &
            \rotatebox[origin=c]{90}{Road marking} &
            \rotatebox[origin=c]{90}{Fabric} &
            \rotatebox[origin=c]{90}{Glass} &
            \rotatebox[origin=c]{90}{Plaster} &
            \rotatebox[origin=c]{90}{Plastic} &
            \rotatebox[origin=c]{90}{Rubber} &
            \rotatebox[origin=c]{90}{Sand} &
            \rotatebox[origin=c]{90}{Gravel} &
            \rotatebox[origin=c]{90}{Ceramic} &
            \rotatebox[origin=c]{90}{Cobblestone} &
            \rotatebox[origin=c]{90}{Brick} &
            \rotatebox[origin=c]{90}{Grass} &
            \rotatebox[origin=c]{90}{Wood} &
            \rotatebox[origin=c]{90}{Leaf} &
            \rotatebox[origin=c]{90}{Water} &
            \rotatebox[origin=c]{90}{Human} &
            \rotatebox[origin=c]{90}{Sky} &
            \rotatebox[origin=c]{90}{Mean} \\ 
            \midrule
            \midrule
            
            RGB &
            83.2 &
            44.2 &
            52.1 &
            70.4 &
            31.0 &
            51.6 &
            1.3 &
            26.2 &
            21.8 &
            65.0 &
            61.8 &
            31.3 &
            72.5 &
            \textbf{45.0} &
            55.4 &
            46.0 &
            74.7 &
            \textbf{56.0} &
            22.7 &
            96.4 &
            50.4 \\
            
            RGB-A &
            86.5 &
            46.5 &
            55.9 &
            \textbf{73.0} &
            35.3 &
            \textbf{56.0} &
            0.8 &
            27.3 &
            27.8 &
            66.2 &
            67.0 &
            28.6 &
            69.6 &
            43.0 &
            57.6 &
            \textbf{49.6} &
            76.4 &
            53.8 &
            8.4 &
            96.6 &
            51.3 \\
            
            RGB-A-D &
            86.0 &
            44.0 &
            55.5 &
            68.1 &
            31.9 &
            54.8 &
            \textbf{2.3} &
            30.0 &
            \textbf{29.7} &
            \textbf{69.4} &
            \textbf{73.7} &
            \textbf{32.2} &
            69.4 &
            41.4 &
            59.2 &
            48.3 &
            76.6 &
            50.6 &
            20.9 &
            \textbf{96.7} &
            52.0 \\ 
        
            RGB-A-D-N &
            \textbf{88.0} &
            \textbf{48.3} &
            \textbf{56.2} &
            72.2 &
            \textbf{35.4} &
            54.9 &
            0.5 &
            \textbf{34.6} &
            29.4 &
            67.2 &
            69.0 &
            29.9 &
            \textbf{73.4} &
            44.7 &
            \textbf{59.5} &
            47.8 &
            \textbf{77.1} &
            50.5 &
            \textbf{26.9} &
            96.6 &
            \textbf{53.1} \\
            \bottomrule
        \end{tabular}
    }
    \label{tab:mcubes-per-class-iou-for-modality-combination}
\end{table*}

\begin{table}[t]
    \centering
    \caption{Comparison of number of parameters in the fusion block and GFLOPs for MCubeS dataset having 4 input modalities with an input shape of $(3 \times 512 \times 512)$ for each modality. Our fusion block shows significantly lower complexity compared to existing methods.}
    \setlength{\tabcolsep}{4pt}
    \begin{tabular}{lcc}
        \toprule
        \multicolumn{1}{c}{Methods} & Fusion Block Parameters (M) & GFLOPs \\
        \midrule
        \midrule
        CMNeXt \cite{zhang2023CMNext}                   & 16.63         & 6.47          \\
        MCubeSNet \cite{Liang2022MCubeS}                & 7.41          & 12.10         \\
        HRFuser \cite{broedermann2023hrfuser}           & \textbf{1.72} & 17.50         \\
        CMX \cite{zhang2023cmx}                         & 16.59         & 6.41          \\
        DDF (Resnet-101) \cite{zhou2021DDF}             & 28.10         & 4.10          \\
        \textbf{MMSFormer (Ours)}                       & 3.23          & \textbf{2.47} \\
        \bottomrule
    \end{tabular}
    \label{tab:gflop-param-comparison}
\end{table}

\subsection{Qualitative analysis of the Predictions}
Apart from quantitative analysis, we also perform qualitative analysis of the predicted segmentation maps. We show material segmentation results predicted by CMNeXt \cite{zhang2023CMNext} model and our proposed MMSFormer model in Figure~\ref{fig:vis-mcubes-sota}. For brevity, we only show RGB images and ground truth material segmentation maps in the illustrations. We show RGB only predictions and all modalities (RGB-A-D-N) predictions for both of the models. As highlighted in the rectangular bounding boxes, our proposed MMSFormer model identifies asphalt, sand and water with greater accuracy than CMNeXt \cite{zhang2023CMNext} model for both RGB only and all modalities (RGB-A-D-N) predictions. 

We also compare our prediction on PST900 \cite{Shivakumar2020PST900} dataset with RTFNet \cite{sun2019rtfnet} and FDCNet \cite{zhao2023FDCNet} on Figure~\ref{fig:vis-pst-sota}. We show the input RGB image, thermal images, ground truth segmentation maps and prediction form the models. As highlighted by the rectangular bounding boxes, our model shows better accuracy in detecting objects with more precise contours compared to the other two methods.

\subsection{Ablation Study on the Fusion Block}
We conducted a number of ablation studies aimed at investigating the contributions of individual components within the fusion block to the overall model performance. The findings, as detailed in Table~\ref{tab:ablation-study-on-fusion-block}, shed light on the critical importance of these components. We used both RGB and infrared modalities of the FMB dataset during training and testing in these experiments. First, we observed that the absence of channel attention in the residual connection had a negative impact, resulting in a reduction in performance by 3.36\%. This indicates that feature calibration along channel dimension plays an important role in capturing and leveraging crucial information effectively. Additionally, while comparing larger convolution kernel sizes ($3\times3$, $7\times 7$, and $11\times 11$) to the originally employed ($3\times3$, $5\times 5$, and $7\times 7$), we noted a decrease in performance by 5.36\%. This result underscores the significance of the carefully chosen convolution kernel sizes within the fusion block.

Furthermore, completely removing the parallel convolutions from the block led to a performance decline of 4.51\%, emphasizing their substantial contribution in capturing multi-scale features and overall model performance. Finally, if we only use the linear fusion layer to fuse the features and remove the parallel convolutions and channel attention from the fusion block, performance drop significantly by 9.25\%. These studies demonstrate that multi-scale feature capturing via parallel convolutions and channel-wise feature calibration using channel attention is extremely important in learning better feature representation and thus crucial to overall model performance. These comprehensive ablation studies collectively underscore the significance of every component within the fusion block, revealing that each module plays a distinct and vital role in achieving the overall performance of our model. 

\subsection{Ablation Study on Different Modality Combinations}
To analyze the contributions of different modalities in the identification of distinct materials, we conducted a series of ablation studies, focusing on per-class IoU for different modality combinations. The insights are summarized in Table~\ref{tab:mcubes-per-class-iou-for-modality-combination}. As we progressively integrate new modalities, performance gradually increases for specific classes, which include grass, leaf, asphalt, cobblestone and plastic classes. Particularly noteworthy is the assistance provided by NIR data in classifying asphalt, concrete, plastic, cobblestone, and human categories, leading to significant performance gains in these classes as NIR was added as an additional modality. 

Conversely, certain classes such as water and brick exhibited a gradual performance decline as we introduced additional modalities. This suggests that RGB data alone suffices for accurately identifying these classes, and the inclusion of more modalities potentially introduces noise or redundancy that negatively impacts performance. Moreover, AoLP appears to be helpful in enhancing the recognition of materials like road markings, glass and wood. Similarly, DoLP improved performance for classes like plaster, rubber, sand, gravel and ceramic. These findings underscore the relationship between different imaging modalities and the unique characteristics of different types of materials, demonstrating that specific modalities excel in detecting particular classes based on their distinctive traits. 

In Figure~\ref{fig:vis-mcubes-modality-combination}, we presents some examples to show how adding different modalities help improve performance of segmentation. We show predictions for RGB, RGB-A, RGB-A-D and RGB-A-D-N inputs from our proposed MMSFormer model. As we add new modalities, the predictions become more accurate as shown in the bounding boxes. The illustrations show that the identification of concrete and gravel becomes more accurate with additional modalities. Figure~\ref{fig:vis-fmb-modality-combination} shows predictions for RGB and RGB-Infrared from FMB dataset. As highlighted by the bounding boxes, adding new modality helps improve performance in detecting building, road and sidewalks. This also illustrates the capability of the fusion block to effectively fuse information from different modality combinations.

\subsection{Computational Complexity of the Fusion Block}
In addition to better performance, our fusion block is also computationally efficient compared to most of the fusion blocks proposed for these datasets. We show a comparison in terms of the number of parameters in the fusion block and GFLOPs for some recent models on Table~\ref{tab:gflop-param-comparison} for MCubeS dataset having 4 input modalities with an input shape of $(3 \times 512 \times 512)$ for each modality. As observed from the table, our proposed fusion strategy demonstrates significantly lower complexity in terms of both the number of parameters and GFLOPs compared to existing methods. HRFuser \cite{broedermann2023hrfuser} has a lower parameter count than ours but it requires more than $7\times$ GFLOPs. Other methods require significantly more parameters $(2.3\times -~8.7\times)$ and GFLOPs $(1.6\times -~7\times)$ compared to our fusion strategy. Our comparison only includes models for which these results are available in the published literature.

\section{Conclusion}
\label{sec:conclusion}
In this paper, we introduce a novel fusion module designed to combine useful information from various modality combinations. We also propose a new model called MMSFormer that integrates the proposed fusion block to accomplish multimodal material and semantic segmentation tasks. Experimental results illustrate the model's capability to efficiently fuse information from different combination of modalities, leading to new state-of-the-art performance on three different datasets. Experiments also show that the fusion block can extract useful information from different modality combinations that helps the model to consistently outperform current state-of-the-art models. Starting from only one input modality, performance increases gradually as we add new modalities. Several ablation studies further highlight how different components of the fusion block contribute to the overall model performance. Ablation studies also reveal that different modalities assist in identifying different types of material classes. However, one limitation of the proposed model is the use of modality specific encoders and the number of encoders grows with number of modalities. Future work will include exploring the possibility and effectiveness of using a shared encoder for all the modalities, investigating and extending the model's capability with other modalities and multimodal tasks. 

\section*{Acknowledgements}
This work is supported in part by AFOSR award FA9550-21-1-0330.
 
\medskip
\bibliography{files/arxiv_single_column/arxiv_single_column}

\begin{thebibliography}{10}

\bibitem{broedermann2023hrfuser}
Tim Broedermann, Christos Sakaridis, Dengxin Dai, and Luc Van~Gool.
\newblock Hrfuser: A multi-resolution sensor fusion architecture for 2d object
  detection.
\newblock In {\em IEEE International Conference on Intelligent Transportation
  Systems (ITSC)}, 2023.

\bibitem{Cai2023DHFNet}
Yuqi Cai, Wujie Zhou, Liting Zhang, Lu~Yu, and Ting Luo.
\newblock Dhfnet: dual-decoding hierarchical fusion network for rgb-thermal
  semantic segmentation.
\newblock {\em The Visual Computer}, pages 1--11, 2023.

\bibitem{chen2021DRConv}
Jin Chen, Xijun Wang, Zichao Guo, Xiangyu Zhang, and Jian Sun.
\newblock Dynamic region-aware convolution.
\newblock In {\em Proceedings of the IEEE/CVF conference on computer vision and
  pattern recognition}, pages 8064--8073, 2021.

\bibitem{chen2018deeplabv3+}
Liang-Chieh Chen, Yukun Zhu, George Papandreou, Florian Schroff, and Hartwig
  Adam.
\newblock Encoder-decoder with atrous separable convolution for semantic image
  segmentation.
\newblock In {\em Proceedings of the European conference on computer vision
  (ECCV)}, pages 801--818, 2018.

\bibitem{chen2020sa-gate}
Xiaokang Chen, Kwan-Yee Lin, Jingbo Wang, Wayne Wu, Chen Qian, Hongsheng Li,
  and Gang Zeng.
\newblock Bi-directional cross-modality feature propagation with
  separation-and-aggregation gate for rgb-d semantic segmentation.
\newblock In {\em European Conference on Computer Vision (ECCV)}, pages
  561--577, 2020.

\bibitem{cheng2022mask2former}
Bowen Cheng, Ishan Misra, Alexander~G Schwing, Alexander Kirillov, and Rohit
  Girdhar.
\newblock Masked-attention mask transformer for universal image segmentation.
\newblock In {\em Proceedings of the IEEE/CVF conference on computer vision and
  pattern recognition}, pages 1290--1299, 2022.

\bibitem{cheng2001color}
Heng-Da Cheng, X\_~H\_ Jiang, Ying Sun, and Jingli Wang.
\newblock Color image segmentation: advances and prospects.
\newblock {\em Pattern recognition}, 34(12):2259--2281, 2001.

\bibitem{deng2021FEANet}
Fuqin Deng, Hua Feng, Mingjian Liang, Hongmin Wang, Yong Yang, Yuan Gao,
  Junfeng Chen, Junjie Hu, Xiyue Guo, and Tin~Lun Lam.
\newblock Feanet: Feature-enhanced attention network for rgb-thermal real-time
  semantic segmentation.
\newblock In {\em 2021 IEEE/RSJ International Conference on Intelligent Robots
  and Systems (IROS)}, page 4467–4473. IEEE Press, 2021.

\bibitem{deng2009ImageNet}
Jia Deng, Wei Dong, Richard Socher, Li-Jia Li, Kai Li, and Li~Fei-Fei.
\newblock Imagenet: A large-scale hierarchical image database.
\newblock In {\em 2009 IEEE Conference on Computer Vision and Pattern
  Recognition}, pages 248--255, 2009.

\bibitem{Dong2022GEBNet}
Shaohua Dong, Wujie Zhou, Xiaohong Qian, and Lu~Yu.
\newblock Gebnet: Graph-enhancement branch network for rgb-t scene parsing.
\newblock {\em IEEE Signal Processing Letters}, 29:2273--2277, 2022.

\bibitem{dong2023EGFNet}
Shaohua Dong, Wujie Zhou, Caie Xu, and Weiqing Yan.
\newblock Egfnet: Edge-aware guidance fusion network for rgb–thermal urban
  scene parsing.
\newblock {\em IEEE Transactions on Intelligent Transportation Systems}, pages
  1--13, 2023.

\bibitem{elharrouss2021panoptic2}
Omar Elharrouss, Somaya~Ali Al-Maadeed, Nandhini Subramanian, Najmath Ottakath,
  Noor Almaadeed, and Yassine Himeur.
\newblock Panoptic segmentation: A review.
\newblock {\em ArXiv}, abs/2111.10250, 2021.

\bibitem{GONG2023GCGLNet}
Tingting Gong, Wujie Zhou, Xiaohong Qian, Jingsheng Lei, and Lu~Yu.
\newblock Global contextually guided lightweight network for rgb-thermal urban
  scene understanding.
\newblock {\em Engineering Applications of Artificial Intelligence},
  117:105510, 2023.

\bibitem{gu2022instanceseg2}
Wenchao Gu, Shuang Bai, and Lingxing Kong.
\newblock A review on 2d instance segmentation based on deep neural networks.
\newblock {\em Image and Vision Computing}, 120:104401, 2022.

\bibitem{guo2018semseg}
Yanming Guo, Yu~Liu, Theodoros Georgiou, and Michael~S Lew.
\newblock A review of semantic segmentation using deep neural networks.
\newblock {\em International journal of multimedia information retrieval},
  7:87--93, 2018.

\bibitem{guo2019deep2}
Zhe Guo, Xiang Li, Heng Huang, Ning Guo, and Quanzheng Li.
\newblock Deep learning-based image segmentation on multimodal medical imaging.
\newblock {\em IEEE Transactions on Radiation and Plasma Medical Sciences},
  3(2):162--169, 2019.

\bibitem{hafiz2020instanceseg}
Abdul~Mueed Hafiz and Ghulam~Mohiuddin Bhat.
\newblock A survey on instance segmentation: state of the art.
\newblock {\em International journal of multimedia information retrieval},
  9(3):171--189, 2020.

\bibitem{hazirbas2017fusenet}
C.~Hazirbas, L.~Ma, C.~Domokos, and D.~Cremers.
\newblock Fusenet: incorporating depth into semantic segmentation via
  fusion-based cnn architecture.
\newblock In {\em Asian Conference on Computer Vision}, November 2016.

\bibitem{hu2019squeezeandexcitation}
Jie Hu, Li~Shen, and Gang Sun.
\newblock Squeeze-and-excitation networks.
\newblock In {\em Proceedings of the IEEE conference on computer vision and
  pattern recognition}, pages 7132--7141, 2018.

\bibitem{hu2019acnet}
Xinxin Hu, Kailun Yang, Lei Fei, and Kaiwei Wang.
\newblock Acnet: Attention based network to exploit complementary features for
  rgbd semantic segmentation.
\newblock In {\em IEEE International Conference on Image Processing (ICIP)},
  pages 1440--1444, 2019.

\bibitem{huang2022Reconet}
Zhanbo Huang, Jinyuan Liu, Xin Fan, Risheng Liu, Wei Zhong, and Zhongxuan Luo.
\newblock Reconet: Recurrent correction network for fast and efficient
  multi-modality image fusion.
\newblock In Shai Avidan, Gabriel Brostow, Moustapha Ciss{\'e}, Giovanni~Maria
  Farinella, and Tal Hassner, editors, {\em Computer Vision -- ECCV 2022},
  pages 539--555, Cham, 2022. Springer Nature Switzerland.

\bibitem{Huang2019CCNet}
Zilong Huang, Xinggang Wang, Lichao Huang, Chang Huang, Yunchao Wei, and Wenyu
  Liu.
\newblock Ccnet: Criss-cross attention for semantic segmentation.
\newblock In {\em IEEE/CVF International Conference on Computer Vision (ICCV)},
  pages 603--612, 2019.

\bibitem{joze2020mmtm}
Hamid Reza~Vaezi Joze, Amirreza Shaban, Michael~L Iuzzolino, and Kazuhito
  Koishida.
\newblock Mmtm: Multimodal transfer module for cnn fusion.
\newblock In {\em Proceedings of the IEEE/CVF conference on computer vision and
  pattern recognition}, pages 13289--13299, 2020.

\bibitem{kirillov2019panoptic}
Alexander Kirillov, Kaiming He, Ross Girshick, Carsten Rother, and Piotr
  Doll{\'a}r.
\newblock Panoptic segmentation.
\newblock In {\em Proceedings of the IEEE/CVF conference on computer vision and
  pattern recognition}, pages 9404--9413, 2019.

\bibitem{Li2023LASNet}
Gongyang Li, Yike Wang, Zhi Liu, Xinpeng Zhang, and Dan Zeng.
\newblock Rgb-t semantic segmentation with location, activation, and
  sharpening.
\newblock {\em IEEE Transactions on Circuits and Systems for Video Technology},
  33(3):1223--1235, 2023.

\bibitem{li2023mseg3d}
J.~Li, H.~Dai, H.~Han, and Y.~Ding.
\newblock Mseg3d: Multi-modal 3d semantic segmentation for autonomous driving.
\newblock In {\em 2023 IEEE/CVF Conference on Computer Vision and Pattern
  Recognition (CVPR)}, pages 21694--21704, Los Alamitos, CA, USA, jun 2023.
  IEEE Computer Society.

\bibitem{li2023RSFNet}
Ping Li, Junjie Chen, Binbin Lin, and Xianghua Xu.
\newblock Residual spatial fusion network for rgb-thermal semantic
  segmentation.
\newblock {\em arXiv:2306.10364}, 2023.

\bibitem{liang2023eaef}
Mingjian Liang, Junjie Hu, Chenyu Bao, Hua Feng, Fuqin Deng, and Tin~Lun Lam.
\newblock Explicit attention-enhanced fusion for rgb-thermal perception tasks.
\newblock {\em {IEEE} Robotics Autom. Lett.}, 8(7):4060--4067, 2023.

\bibitem{Liang2022MCubeS}
Yupeng Liang, Ryosuke Wakaki, Shohei Nobuhara, and Ko~Nishino.
\newblock Multimodal material segmentation.
\newblock In {\em Proceedings of the IEEE/CVF Conference on Computer Vision and
  Pattern Recognition (CVPR)}, pages 19800--19808, June 2022.

\bibitem{liu2020EfficientFCN}
Jianbo Liu, Junjun He, Jiawei Zhang, Jimmy~S. Ren, and Hongsheng Li.
\newblock Efficientfcn: Holistically-guided decoding for semantic segmentation.
\newblock In {\em Computer Vision -- ECCV 2020}, pages 1--17. Springer
  International Publishing, 2020.

\bibitem{Liu2022GCNet}
Jinfu Liu, Wujie Zhou, Yueli Cui, Lu~Yu, and Ting Luo.
\newblock Gcnet: Grid-like context-aware network for rgb-thermal semantic
  segmentation.
\newblock {\em Neurocomput.}, 506(C):60–67, sep 2022.

\bibitem{liu2022tardal}
Jinyuan Liu, Xin Fan, Zhanbo Huang, Guanyao Wu, Risheng Liu, Wei Zhong, and
  Zhongxuan Luo.
\newblock Target-aware dual adversarial learning and a multi-scenario
  multi-modality benchmark to fuse infrared and visible for object detection.
\newblock In {\em Proceedings of the IEEE/CVF Conference on Computer Vision and
  Pattern Recognition}, pages 5802--5811, 2022.

\bibitem{liu2023segmif}
Jinyuan Liu, Zhu Liu, Guanyao Wu, Long Ma, Risheng Liu, Wei Zhong, Zhongxuan
  Luo, and Xin Fan.
\newblock Multi-interactive feature learning and a full-time multi-modality
  benchmark for image fusion and segmentation.
\newblock In {\em International Conference on Computer Vision}, 2023.

\bibitem{long2015FCN}
Jonathan Long, Evan Shelhamer, and Trevor Darrell.
\newblock Fully convolutional networks for semantic segmentation.
\newblock In {\em Proceedings of the IEEE conference on computer vision and
  pattern recognition}, pages 3431--3440, 2015.

\bibitem{loshchilov2017adamw}
Ilya Loshchilov and Frank Hutter.
\newblock Decoupled weight decay regularization.
\newblock In {\em International Conference on Learning Representations}, 2019.

\bibitem{minaee2021image}
Shervin Minaee, Yuri Boykov, Fatih Porikli, Antonio Plaza, Nasser Kehtarnavaz,
  and Demetri Terzopoulos.
\newblock Image segmentation using deep learning: A survey.
\newblock {\em IEEE transactions on pattern analysis and machine intelligence},
  44(7):3523--3542, 2021.

\bibitem{minaee2021imagesegsinglemod}
Shervin Minaee, Yuri Boykov, Fatih Porikli, Antonio Plaza, Nasser Kehtarnavaz,
  and Demetri Terzopoulos.
\newblock Image segmentation using deep learning: A survey.
\newblock {\em IEEE transactions on pattern analysis and machine intelligence},
  44(7):3523--3542, 2021.

\bibitem{prakash2021TransFuser}
Aditya Prakash, Kashyap Chitta, and Andreas Geiger.
\newblock Multi-modal fusion transformer for end-to-end autonomous driving.
\newblock In {\em Proceedings of the IEEE/CVF Conference on Computer Vision and
  Pattern Recognition}, pages 7077--7087, 2021.

\bibitem{ronneberger2015unet}
Olaf Ronneberger, Philipp Fischer, and Thomas Brox.
\newblock U-net: Convolutional networks for biomedical image segmentation.
\newblock In {\em Medical Image Computing and Computer-Assisted
  Intervention--MICCAI 2015: 18th International Conference, Munich, Germany,
  October 5-9, 2015, Proceedings, Part III 18}, pages 234--241. Springer, 2015.

\bibitem{Shivakumar2020PST900}
Shreyas~S. Shivakumar, Neil Rodrigues, Alex Zhou, Ian~D. Miller, Vijay Kumar,
  and Camillo~J. Taylor.
\newblock Pst900: Rgb-thermal calibration, dataset and segmentation network.
\newblock In {\em 2020 IEEE International Conference on Robotics and Automation
  (ICRA)}, pages 9441--9447, 2020.

\bibitem{sun2019rtfnet}
Yuxiang Sun, Weixun Zuo, and Ming Liu.
\newblock {RTFNet: RGB-Thermal Fusion Network for Semantic Segmentation of
  Urban Scenes}.
\newblock {\em {IEEE Robotics and Automation Letters}}, 4(3):2576--2583, July
  2019.

\bibitem{upchurch2022dense}
Paul Upchurch and Ransen Niu.
\newblock A dense material segmentation dataset for indoor and outdoor scene
  parsing.
\newblock In {\em European Conference on Computer Vision}, pages 450--466.
  Springer, 2022.

\bibitem{wang2022CGFNet}
Jie Wang, Kechen Song, Yanqi Bao, Liming Huang, and Yunhui Yan.
\newblock Cgfnet: Cross-guided fusion network for rgb-t salient object
  detection.
\newblock {\em IEEE Transactions on Circuits and Systems for Video Technology},
  32(5):2949--2961, 2022.

\bibitem{wang2018semseg2}
Panqu Wang, Pengfei Chen, Ye~Yuan, Ding Liu, Zehua Huang, Xiaodi Hou, and
  Garrison Cottrell.
\newblock Understanding convolution for semantic segmentation.
\newblock In {\em 2018 IEEE winter conference on applications of computer
  vision (WACV)}, pages 1451--1460. Ieee, 2018.

\bibitem{wang2021pvt}
Wenhai Wang, Enze Xie, Xiang Li, Deng-Ping Fan, Kaitao Song, Ding Liang, Tong
  Lu, Ping Luo, and Ling Shao.
\newblock Pyramid vision transformer: A versatile backbone for dense prediction
  without convolutions.
\newblock In {\em in IEEE/CVF international conference on computer vision},
  pages 568--578, 2021.

\bibitem{woo2018cbam}
Sanghyun Woo, Jongchan Park, Joon-Young Lee, and In~So Kweon.
\newblock Cbam: Convolutional block attention module.
\newblock In {\em Proceedings of the European conference on computer vision
  (ECCV)}, pages 3--19, 2018.

\bibitem{wu2019fastfcn}
Huikai Wu, Junge Zhang, Kaiqi Huang, Kongming Liang, and Yizhou Yu.
\newblock Fastfcn: Rethinking dilated convolution in the backbone for semantic
  segmentation.
\newblock {\em arXiv preprint arXiv:1903.11816}, 2019.

\bibitem{xie2021segformer}
Enze Xie, Wenhai Wang, Zhiding Yu, Anima Anandkumar, Jose~M Alvarez, and Ping
  Luo.
\newblock Segformer: Simple and efficient design for semantic segmentation with
  transformers.
\newblock In {\em Neural Information Processing Systems (NeurIPS)}, 2021.

\bibitem{xu2022U2Fusion}
Han Xu, Jiayi Ma, Junjun Jiang, Xiaojie Guo, and Haibin Ling.
\newblock U2fusion: A unified unsupervised image fusion network.
\newblock {\em IEEE Transactions on Pattern Analysis and Machine Intelligence},
  44(1):502--518, 2022.

\bibitem{zhang2023cmx}
Jiaming Zhang, Huayao Liu, Kailun Yang, Xinxin Hu, Ruiping Liu, and Rainer
  Stiefelhagen.
\newblock Cmx: Cross-modal fusion for rgb-x semantic segmentation with
  transformers.
\newblock {\em IEEE Transactions on Intelligent Transportation Systems}, 2023.

\bibitem{zhang2023CMNext}
Jiaming Zhang, Ruiping Liu, Hao Shi, Kailun Yang, Simon Rei{\ss}, Kunyu Peng,
  Haodong Fu, Kaiwei Wang, and Rainer Stiefelhagen.
\newblock Delivering arbitrary-modal semantic segmentation.
\newblock In {\em Proceedings of the IEEE/CVF Conference on Computer Vision and
  Pattern Recognition}, pages 1136--1147, 2023.

\bibitem{zhang2021deep}
Yifei Zhang, D{\'e}sir{\'e} Sidib{\'e}, Olivier Morel, and Fabrice
  M{\'e}riaudeau.
\newblock Deep multimodal fusion for semantic image segmentation: A survey.
\newblock {\em Image and Vision Computing}, 105:104042, 2021.

\bibitem{zhang2021imagesegmm}
Yifei Zhang, D{\'e}sir{\'e} Sidib{\'e}, Olivier Morel, and Fabrice
  M{\'e}riaudeau.
\newblock Deep multimodal fusion for semantic image segmentation: A survey.
\newblock {\em Image and Vision Computing}, 105:104042, 2021.

\bibitem{zhao2017pspnet}
Hengshuang Zhao, Jianping Shi, Xiaojuan Qi, Xiaogang Wang, and Jiaya Jia.
\newblock Pyramid scene parsing network.
\newblock In {\em Proceedings of the IEEE conference on computer vision and
  pattern recognition}, pages 2881--2890, 2017.

\bibitem{zhao2021lifseg}
Lin Zhao, Hui Zhou, Xinge Zhu, Xiao Song, Hongsheng Li, and Wenbing Tao.
\newblock Lif-seg: Lidar and camera image fusion for 3d lidar semantic
  segmentation.
\newblock {\em IEEE Transactions on Multimedia}, pages 1--11, 2023.

\bibitem{Zhao2023MDRNet+}
Shenlu Zhao, Yichen Liu, Qiang Jiao, Qiang Zhang, and Jungong Han.
\newblock Mitigating modality discrepancies for rgb-t semantic segmentation.
\newblock {\em IEEE Transactions on Neural Networks and Learning Systems},
  pages 1--15, 2023.

\bibitem{zhao2023FDCNet}
Shenlu Zhao and Qiang Zhang.
\newblock A feature divide-and-conquer network for rgb-t semantic segmentation.
\newblock {\em IEEE Transactions on Circuits and Systems for Video Technology},
  33(6):2892--2905, 2023.

\bibitem{Zhao2020DIDFuse}
Zixiang Zhao, Shuang Xu, Chunxia Zhang, Junmin Liu, Jiangshe Zhang, and Pengfei
  Li.
\newblock Didfuse: Deep image decomposition for infrared and visible image
  fusion.
\newblock In {\em {IJCAI}}, pages 970--976. ijcai.org, 2020.

\bibitem{zhou2021DDF}
Jingkai Zhou, Varun Jampani, Zhixiong Pi, Qiong Liu, and Ming-Hsuan Yang.
\newblock Decoupled dynamic filter networks.
\newblock In {\em Proceedings of the IEEE/CVF Conference on Computer Vision and
  Pattern Recognition}, pages 6647--6656, 2021.

\bibitem{Zhou2023CACFNet}
Wujie Zhou, Shaohua Dong, Meixin Fang, and Lu~Yu.
\newblock Cacfnet: Cross-modal attention cascaded fusion network for rgb-t
  urban scene parsing.
\newblock {\em IEEE Transactions on Intelligent Vehicles}, pages 1--10, 2023.

\bibitem{zhou2023MTANet}
Wujie Zhou, Shaohua Dong, Jingsheng Lei, and Lu~Yu.
\newblock Mtanet: Multitask-aware network with hierarchical multimodal fusion
  for rgb-t urban scene understanding.
\newblock {\em IEEE Transactions on Intelligent Vehicles}, 8(1):48--58, 2023.

\bibitem{zhou2022EGFNet1}
Wujie Zhou, Shaohua Dong, Caie Xu, and Yaguan Qian.
\newblock Edge-aware guidance fusion network for rgb--thermal scene parsing.
\newblock In {\em Proceedings of the AAAI Conference on Artificial
  Intelligence}, volume~36, pages 3571--3579, 2022.

\bibitem{Zhou2022MFFENet}
Wujie Zhou, Xinyang Lin, Jingsheng Lei, Lu~Yu, and Jenq-Neng Hwang.
\newblock Mffenet: Multiscale feature fusion and enhancement network for
  rgb–thermal urban road scene parsing.
\newblock {\em IEEE Transactions on Multimedia}, 24:2526--2538, 2022.

\bibitem{zhou2021GMNet}
Wujie Zhou, Jinfu Liu, Jingsheng Lei, Lu~Yu, and Jenq-Neng Hwang.
\newblock Gmnet: Graded-feature multilabel-learning network for rgb-thermal
  urban scene semantic segmentation.
\newblock {\em IEEE Transactions on Image Processing}, 30:7790--7802, 2021.

\end{thebibliography}
\bibliographystyle{plain}

\end{document}